\documentclass[10pt,twocolumn,letterpaper]{article}

\usepackage{cvpr}              
\usepackage{wrapfig}
\usepackage{multirow}
\usepackage{makecell}
\usepackage{bbding}
\usepackage{color}

\usepackage{xcolor}

\definecolor{cvprblue}{rgb}{0.21,0.49,0.74}
\usepackage[pagebackref,breaklinks,colorlinks,citecolor=cvprblue]{hyperref}

\def\paperID{3945} 
\def\confName{CVPR}
\def\confYear{2024}

\title{A Unified Framework for Human-centric Point Cloud Video Understanding}

\author{\textbf{
Yiteng Xu\textsuperscript{\rm 1},
Kecheng Ye\textsuperscript{\rm 1},
Xiao Han\textsuperscript{\rm 1},
Yiming Ren\textsuperscript{\rm 1},
Xinge Zhu\textsuperscript{\rm 2},
Yuexin Ma$^{1,}$\thanks{Corresponding author. This work was supported by NSFC (No.62206173), Shanghai Sailing Program (No.22YF1428700), MoE Key Laboratory of Intelligent Perception and Human-Machine Collaboration (ShanghaiTech University), Shanghai Frontiers Science Center of Human-centered Artificial Intelligence (ShangHAI), Shanghai Engineering Research Center of Intelligent Vision and Imaging.}
}\\ 
$^{1}$ ShanghaiTech University
$^{2}$ The Chinese University of Hong Kong \\
{\tt\small \{xuyt2023,mayuexin\}@shanghaitech.edu.cn}}

\begin{document}
\maketitle
\begin{abstract}
Human-centric Point Cloud Video Understanding (PVU) is an emerging field focused on extracting and interpreting human-related features from sequences of human point clouds, further advancing downstream human-centric tasks and applications. Previous works usually focus on tackling one specific task and rely on huge labeled data, which has poor generalization capability. Considering that human has specific characteristics, including the structural semantics of human body and the dynamics of human motions, we propose a unified framework to make full use of the prior knowledge and explore the inherent features in the data itself for generalized human-centric point cloud video understanding. Extensive experiments demonstrate that our method achieves state-of-the-art performance on various human-related tasks, including action recognition and 3D pose estimation. \textbf{All datasets and code will be released soon.}

\end{abstract}    
\section{Introduction}
\label{sec:intro}

Human-centric point cloud video understanding (PVU) is a burgeoning field focused on discerning, interpreting, and quantifying human-related information within sequences of human point clouds. This area has witnessed a surge in attention in recent years, particularly applied in LiDAR captured large-scale unconstrained scenarios~\cite{xu2023human,dai2023sloper4d,yan2023cimi4d,han2022licamgait}. Its significance lies in its critical role in facilitating various downstream tasks, including human action recognition~\cite{xu2023human}, 3D pose estimation~\cite{cong2023weakly}, motion capture~\cite{ren2023lidar,li2022lidarcap}, etc. These advancements hold the potential to further drive progress in real-world applications, such as intelligent surveillance, assistive robots, human-robot collaboration, etc.

Current methods~\cite{xu2023human,ren2023lidar,li2022lidarcap} usually rely on extensive labeled data for supervision and employ generic point cloud-based feature extraction backbones~\cite{qi2017pointnet,qi2017pointnet++,ma2022rethinking,guo2021pct}. Nevertheless, obtaining the necessary data and annotations for 4D human-centric point cloud videos proves to be a challenging and expensive endeavor. Furthermore, fully supervised techniques tend to exhibit overfitting issues when applied to specific datasets or tasks, resulting in limited generalization capabilities. Additionally, the existing feature extraction networks are ill-suited for human-centric data, as they fail to account for human-specific characteristics. Hence, within the domain of human-centric PVU, the significance of self-supervised learning becomes evident in enhancing algorithmic generalization. Simultaneously, the development of a human-specific feature extractor that uses prior human-related knowledge holds great promise in bolstering the effectiveness of methods for downstream tasks.
\begin{figure}[t]
    \centering  
    \includegraphics[width=0.5\textwidth]
    {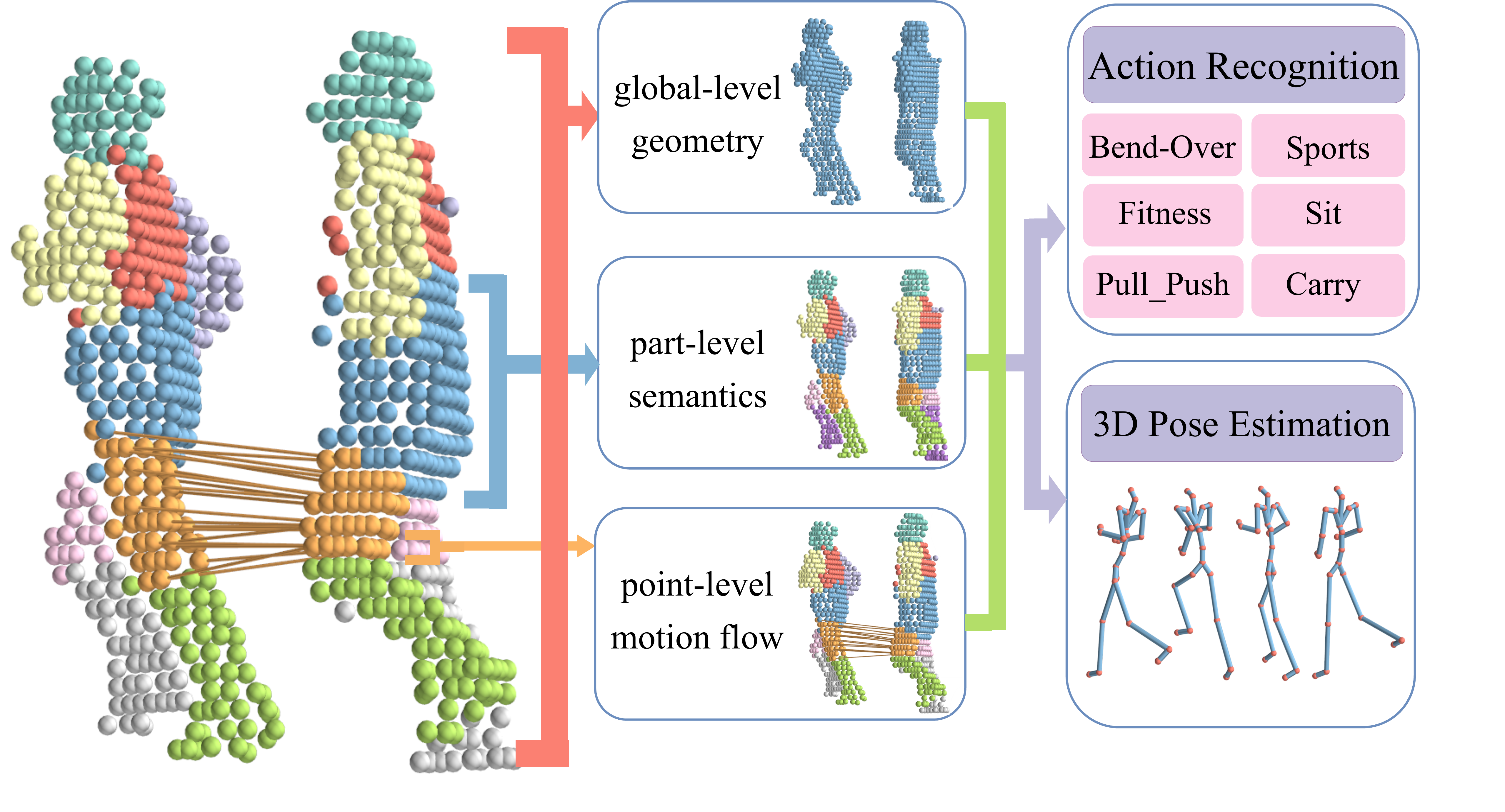} 
    \caption{UniPVU-Human extracts human-related prior knowledge at global level, part level, and point level to facilitate subsequent geometric and dynamic representation learning, finally cater to a range of downstream human-centric tasks, such as action recognition, 3D pose estimation, etc.}
    \label{fig:overview}
 \end{figure}

Actually, self-supervised learning~\cite{wang2021self,zhang2023complete} for PVU has made great progress. Some approaches~\cite{sheng2023contrastive,sheng2023point,shen2023pointcmp} leverage contrastive learning techniques to capture essential spatio-temporal features within dynamic point clouds. Nevertheless, due to the inherent challenges posed by irregular point distributions stemming from varying capture distances, occlusions, and noise, the construction of high-quality positive and negative samples remains a nontrivial task, thereby making the optimization process difficult. The latest work~\cite{shen2023masked} exploits mask prediction for point cloud video self-learning by dividing sequential point clouds into tubes for masking and recovering. However, this method flattens all tubes for feature learning, inadvertently compromising the semantic and dynamic consistency in 4D videos. Moreover, all these methods are not tailored specifically to address human-centric PVU.

Given the importance of human-centric tasks, the imperative need arises to establish a unified framework for the understanding of human point cloud videos. Notably, no specific solutions to this challenge have been identified to date. In this paper, we approach the problem by addressing two fundamental questions: \textit{first, what human-related prior knowledge can be extracted, and second, how can the knowledge be harnessed to enhance human-centric representation learning}?

Considering the inherent structure of the human body, characterized by fixed components such as torso, head, arms, and legs, as well as the distinctive dynamic traits exhibited during human motion, we exploit both \textbf{the structural semantics of human body and human motion dynamics} to facilitate the acquisition of human-specific features from sequences of point clouds. In particular, we create two large-scale point-cloud-based datasets and corresponding pre-trained networks for body segmentation and motion flow estimation, respectively, so that human prior knowledge can be learned in advance and assist subsequent representation learning. 
Furthermore, within our framework, we introduce two innovative stages tailored to maximize the utility of this prior knowledge. The first one, termed \textbf{semantic-guided spatio-temporal representation self-learning}, incorporates a body-part-based mask prediction mechanism designed to facilitate the acquisition of geometric and dynamic representations of humans in the absence of annotations. Building upon this foundation, the following stage, \textbf{hierarchical feature enhanced fine-tuning}, integrates and adapts global-level, part-level, and point-level point cloud features to cater to a range of downstream tasks. In this way, our approach, named \textbf{UniPVU-Human}, serves as a comprehensive exploration of human prior knowledge, furnishing a unified framework for the effective learning of human-centric representations.

To evaluate the effectiveness of our method, we conduct extensive experiments on two popular LiDAR-point-cloud-based datasets~\cite{xu2023human,ren2023lidar}, focusing on human action recognition and human pose estimation, respectively. Our method achieves state-of-the-art performance on both tasks. Detailed ablation studies are also provided to verify each stage and technical design in our framework.

To summarize, our contributions are as follows:
\begin{enumerate}
    \item We propose UniPVU-Human, the first unified framework for human-centric point cloud video understanding, which is significant for vast downstream applications.
    \item Containing two novel stages, including semantic-guided spatio-temporal representation self-learning and hierarchical feature enhanced fine-tuning, our method fully takes advantage of prior knowledge of humans for effective and robust human-centric representation learning.
    \item Our method achieves state-of-the-art performance on open datasets for various human-centric tasks.
    
\end{enumerate}

\section{Related Work}
\label{sec:related_work}
\begin{figure*}[t]
    \centering
    \includegraphics[width=2.1\columnwidth]{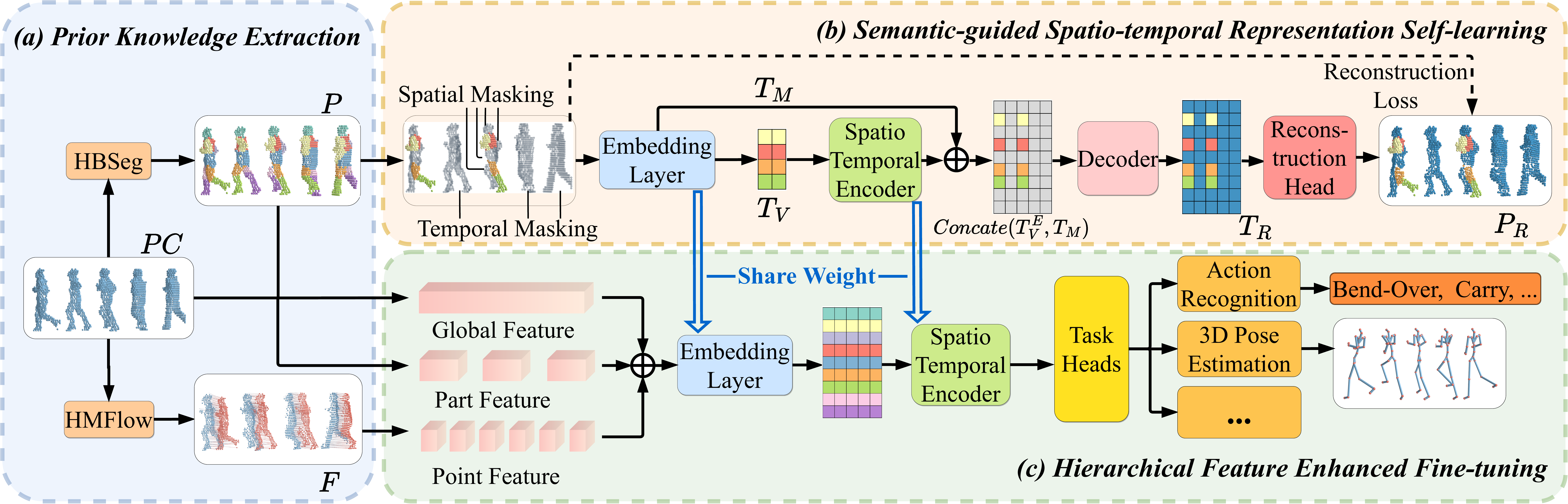}
    \caption{The main pipeline of UniPVU-Human, which can be divided into three stages, including (a) Prior Knowledge Extraction, (b) Semantic-Guided Spatio-temporal Representation Self-learning, and (c) Hierarchical Feature Enhanced Fine-tuning. First, the pre-trained HBSeg and HMFlow are used to provide geometric and dynamic information, including body part segmentation results and point-wise motion flow. Then, our self-learning stage incorporates a body-part-based mask prediction mechanism designed to facilitate the acquisition of geometric and dynamic representations of humans in the absence of annotations. Finally, we integrate global-level, part-level, and point-level features to boost the knowledge transfer to downstream tasks in the fine-tuning stage. } 
    \label{fig:main_figure}
    \vspace{-1ex}
\end{figure*}
\subsection{Feature Learning for Point Clouds}
Point cloud is an important representation for 3D scenes and objects, and tremendous efforts have been made~\cite{ma2022rethinking,qian2022pointnext} for extracting valuable features from point clouds. PointNet~\cite{qi2017pointnet} is to learn a spatial encoding of each point and then aggregate all individual point features to a global point cloud signature\cite{qi2017pointnet++}.
PointNet++\cite{qi2017pointnet++} further introduces a hierarchical feature learning paradigm to capture the local geometric structures recursively. 
PointNeXt~\cite{qian2022pointnext} revisits PointNet++\cite{qi2017pointnet++} with improved
training and scaling strategies.
PCT~\cite{guo2021pct} and PointTransformer~\cite{zhao2021point} apply attention-based mechanism~\cite{vaswani2017attention} to point cloud representations.
Subsequently, many methods~\cite{choy20194d,liu2019meteornet,choy20194d,fan2022pstnet} extend them to process dynamic point cloud videos for more extensive applications.  
P4Transformer~\cite{fan2021point} and PST-Transformer~\cite{fan2022point} use transformers among all local 4D tubes' features to capture long relationships. With the development of autonomous driving, some works~\cite{voxelnet,zhu2021cylindrical,centerformer,centerpoint} propose voxel-based or voxel-point-based feature extractors to process LiDAR point clouds for high efficiency.
However, all these methods are not specifically designed for human dynamic point clouds, lacking the consideration of human-specific characteristics. 

\subsection{LiDAR-based Human-centric Understanding}
Recently, the understanding of human-centric point cloud videos, which are captured by LiDARs in large-scale scenes, has become an emerging field with a lot of new datasets and benchmarks for various human-centric tasks. 
LiDARCap~\cite{li2022lidarcap} contributes an in-the-wild human motion dataset and proposes a LiDAR point cloud video-based motion capture framework. Subsequent works, LIP~\cite{ren2023lidar}, explores the feature fusion of different visual sensors to address 3D pose estimation task based on point clouds and images. 
Recently, HuCenLife~\cite{xu2023human} proposes a huge human-centric dataset with diverse daily-life scenarios and rich human activities, and provides baselines for human perception, action recognition, motion prediction, etc. However, all these approaches follow previous generic backbones to extract features from sequence human point clouds without making use of human prior knowledge. Moreover, they are all supervised methods, causing unsatisfactory results when generalizing to other datasets or tasks.

\subsection{Self-supervised Learning for Point Clouds}
To understand the point cloud representation from data itself instead of supervision by manual annotations, some methods improve the generalization capability via self-learning in the pre-training stage. There exist many methods~\cite{xie2020pointcontrast,yu2022point,pang2022masked,zhang2022point,chen2023pointgpt} learning the geometric representation from static point cloud in a self-supervised manner. Recently, more and more self-learning methods~\cite{wang2021self,zhang2023complete} are proposed to learn the spatio-temporal representations of point cloud videos. Some approaches~\cite{sheng2023contrastive,sheng2023point,shen2023pointcmp} adopt contrastive learning spatially and among frames to learn inherent geometric and dynamic features. However, LiDAR point clouds are sparsity-varying across different capture distances, are usually incomplete due to occlusions, and contain undesired noises, making these methods unstable due to low-quality positive and negative pairs. In the recent study~\cite{shen2023masked}, mask prediction~\cite{he2022masked} is employed for self-learning in point cloud videos by segmenting sequential point clouds into tubes. However, it flattens all tubes during feature learning, inadvertently affecting the semantic and dynamic consistency in 4D videos. Additionally, all these methods lack customization for the specific requirements of human-centric point cloud video understanding.

\label{sec:method}

\section{Method}
The whole architecture of our method is presented in Figure.~\ref{fig:main_figure}. A point cloud video sequence for human instances is denoted as $PC \in 
 \mathcal{R}^{L \times N \times D}$, where $L$ is the sequence length, $N$ is number of points in each frame, and $D$ is the dimension of each point. 
To exploit both the structural semantics of human body and human motion to facilitate the acquisition of human-specific features from sequences of point clouds, we create two large-scale point-cloud-based datasets and corresponding pre-trained networks for body segmentation and motion flow estimation in \textbf{Prior Knowledge Extraction}.
Besides, to learn the essential geometric and dynamic representations of humans from data itself, we explore the spatial and temporal relationships of structural semantics in human body by applying spatio-temporal modeling upon the embedding of body parts in the stage of \textbf{Semantic-guided Spatio-temporal Representation Self-learning}.
Building upon this foundation, we use extracted multiple levels of human-related prior knowledge to benefit various downstream tasks by \textbf{Hierarchical Feature Enhanced Fine-tuning}, which integrates global-level, part-level, and point-level point cloud features.

\subsection{Prior Knowledge Extraction}
Different from the movements of rigid objects such as vehicles in traffic scenarios, the motion of non-rigid humans is more complicated, for it contains not only global rotations and translations but also local rotations and translations, such as relative motions among joints, making capturing human motion features more challenging. The key to addressing this problem is to model the human motion in more fine-grained levels, including body part level and point-wise level. Therefore, we build the Human Body Segmentation (HBSeg) and Human Motion Flow (HMFlow) networks to provide more fine-grained geometric and motion information about human body, which can serve as prior knowledge to facilitate following human-centric representation learning.
\subsubsection{Human Body Segmentation (HBSeg)}
\begin{figure}[h]
    \centering  
    \includegraphics[width=0.5\textwidth]
    {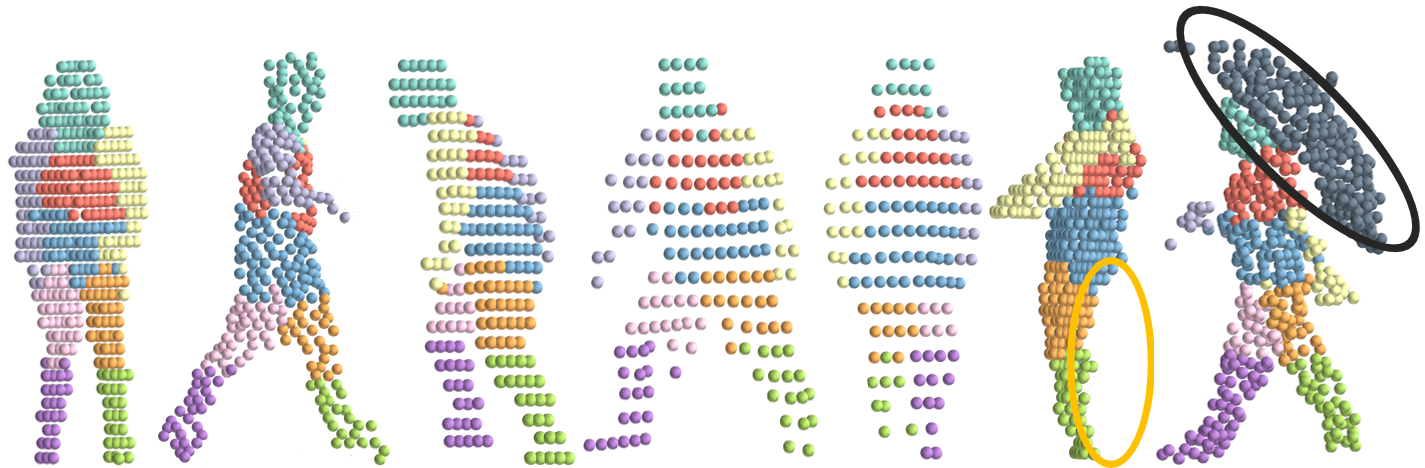} 
    \caption{Visualization results of HBSeg on HuCenLife~\cite{xu2023human}.
    We show cases with different densities of LiDAR point cloud, occlusion (yellow circle), and noise (black circle). HBSeg has robust performance even merely trained on our synthesized dataset.}

    \label{fig:partseg_6.png}
 \end{figure}

\label{subsec:HBPSegModule}
To fully utilize the structural semantics of human body, we deconstruct human into fine-grained body parts by HBSeg, pre-trained on our synthetic dataset, which is constructed by employing a simulated LiDAR model~\cite{ren2023lidar} to scan the surfaces of 3D human meshes from the AMASS dataset~\cite{mahmood2019amass} at various perspectives and distances, meanwhile introducing random occlusions and noise to minimize the distribution gap between synthetic data and real data. We define 9 parts of human body and generate annotations by attaching the body part label of the nearest SMPL~\cite{loper2023smpl} mesh vertex to the synthetic LiDAR Point. More details are in Section.~\ref{subsubsec:HBPSegData} and supplementary material.

We adopt the PointNeXt-L~\cite{qian2022pointnext} as the main body network of HBSeg. 
After training on synthetic data, we apply the pre-trained HBSeg on real data to get the part segmentation labels $S \in \mathcal{R}^{L \times N \times 1}$. As Figure.~\ref{fig:partseg_6.png} shows, HBSeg performs stable on real-life data with changing sparsity and works well even for occlusion and noise cases, mainly due to our efforts on generating realistic synthetic data.

\subsubsection{Human Motion Flow Estimation (HMFlow)}

\begin{figure}[h]
    \centering  
    \includegraphics[width=0.46\textwidth]{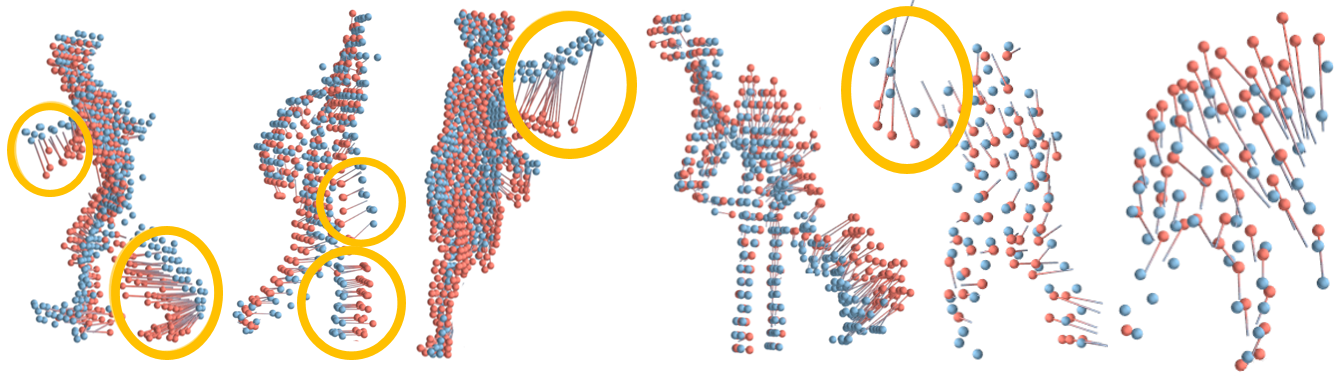}  
    \caption{Visualization results of HMFlow on HuCenLife~\cite{xu2023human}. We present several cases from near to far relative to the LiDAR sensor. HMFlow has good capability of estimating point flow even for the parts with significant movements (yellow circle), which can provide explicit features of human dynamics.}

    \label{fig:vis_HMFlow}
 \end{figure}

\label{subsec:HMFlowModule}
We pre-train the HMFlow on our synthetic dataset to provide point-wise motion information, which can benefit the feature enhancement in the fine-tuning stage.
Similar to Section.~\ref{subsec:HBPSegModule}, we associate each synthetic LiDAR point to its nearest SMPL vertex. Therefore, we are able to establish the correspondence between synthetic LiDAR points across different frames by using SMPL vertices indices, so that we can obtain motion flow ground truth for training our HMFlow. More details are in Section.~\ref{subsubsec:HMFlowData} and supplementary material.

We employ FLOT~\cite{puy2020flot} as the human motion flow estimator, which casts the task of scene flow estimation as finding soft correspondences on a pair of point clouds via solving an optimal transport problem~\cite{li2023deep}. When testing on the real data, we input adjacent LiDAR point clouds $PC_t$ and $PC_{t+1} \in \mathcal{R}^{L \times N \times D}$ into the pre-trained HMFlow to obtain human motion flow vectors $F \in \mathcal{R}^{L \times N \times D'}$ for each LiDAR point in the t-th frame. As Figure.~\ref{fig:vis_HMFlow} demonstrates, HMFlow can generate reasonable prediction for point-wise motion flow on real data even without annotations, which is valuable to provide explicit priors for human dynamics.


\subsection{Semantic-guided Spatio-temporal Representation Self-learning}
\label{subsec:STMaskRecon}
Due to spatial irregularities and temporal redundancies, annotating dynamic point cloud videos is labor-intensive and error-prone. Moreover, fully-supervised methods usually overfit to manual annotations in specific domains, struggling to capture the underlying patterns of new data~\cite{zeng2023self}. This limitation leads to a restricted ability to generalize to other tasks or datasets. Additionally, existing feature extraction networks employ generic point-cloud-based backbones that are ill-suited for human-centric data, as they do not account for human-specific characteristics.
Based on structure semantics of human bodies obtained in previous stage, we propose a module, named Semantic-guided Spatio-temporal Representation Self-learning, which mines essential geometric and motion features from human point cloud video data itself by masking and predicting body part patches to enhance the generalization ability of the model to benefit a variety of downstream tasks.

As Figure.~\ref{fig:main_figure} shows, We first mask some of the body-part tokens in temporal and spatial dimensions. After extracting the token embedding of part patches, only visible tokens will be input to the STEncoder to extract latent representation, which will be decoded with masked tokens together to reconstruct the 3D Cartesian coordinates of masked part patches. Details are as follows.

\subsubsection{Spatio-temporal Masking Strategy}
In temporal dimension, we mask all part patches in some random frames and reconstruct the masked tokens, encouraging STEncoder networks to estimate the part motion over a long period.
In spatial dimension, we randomly mask some part patches in the remaining frames after temporal masking, making the STEncoder network estimate the spatial geometric features of the entire human based on the visible tokens.

We apply temporal masking first, and then spatial masking. Given part patches $P \in \mathcal{R}^{L \times M \times N' \times D}$, we adopt a temporal mask ratio $r_t$ and spatial mask ratio $r_s$, respectively. Firstly, all part patches in $r_tL$ frames are masked, named as temporal masked patches $P_M^t \in \mathcal{R}^{r_tL \times M \times N' \times D}$, which will be used as the reconstruction ground truth of temporal masking. For every frame of the remaining $(1-r_t)L$ visible frames, $r_sM$ part patches are masked randomly, hence spatial masked patches $P_M^s \in \mathcal{R}^{(1-r_t)L \times r_sM \times N' \times D}$ will be used as the reconstruction ground truth of the spatial masking.

\subsubsection{Embedding Layer}
\label{subsubsec:embedding}
For visible part patches, we use Mini-PointNet~\cite{qi2017pointnet} as a tokenizer to obtain visible part tokens embedding $T_V$ from visible part patches $P_V$. 
\begin{equation}
    \begin{aligned}
    T_V = Tokenizer(P_V),
    \end{aligned}
\end{equation}
where $T_V \in \mathcal{R}^{(1-r_t)L \times (1-r_s)M \times C}$, $P_V \in \mathcal{R}^{(1-r_t)L \times (1-r_s)M \times N' \times D}$. $C$ is the channel dimension of part tokens. 
For every masked part patch, we replace it with a share-weighted learnable masked part token $T_M$, which will be concatenated with the output of STEncoder and processed by the decoder together.

Learnable spatial positional encoding and temporal positional encoding will also be added to the input of every transformer layer in the STEncoder and decoder.

\subsubsection{Spatio-temporal Encoder (STEncoder)}

To fully utilize the inherent structure of the human body, characterized by fixed components such as torso, head, arms, and legs, as well as the distinctive dynamic traits exhibited during human motion, our STEncoder applies spatial modeling and temporal modeling on body part tokens, respectively.
The STEncoder learns to extract the high-level latent geometric and motion features of humans from only $T_V$, which are input to STEncoder to get enhanced visible part tokens $T_V^E$.

For the network design of STEncoder, we interlace multiple spatial transformer~\cite{vaswani2017attention} layers and temporal transformer layers to extract the spatial geometry feature and temporal motion feature, respectively. For each spatial transformer layer, we apply self-attention~\cite{vaswani2017attention} among all visible parts tokens in every frame:
\begin{equation}
    \begin{aligned}
    V^{s'}_T = SpatialTransformer(V^{s}_T),
    \end{aligned}
\end{equation}
where $V^{s}_T, V^{s'}_T \in \mathcal{R}^{(1-r_s)M \times C}$ are the visible part tokens in a frame. For each temporal transformer layer, we apply self-attention for every visible part token among all L frames:
\begin{equation}
    \begin{aligned}
    V^{t'}_T = TemporalTransformer(V^{t}_T),
    \end{aligned}
\end{equation}
where $V^{t}_T, V^{t'}_T \in \mathcal{R}^{(1-r_t)L\times C}$ are the visible part tokens of a body part in $L$ frames.

\subsubsection{Mask Reconstruction}
Our decoder is similar to the STEncoder with fewer layers. We take the enhanced visible part tokens $T_V^E$ as well as masked part token $T_M$ as the input of the decoder. 
\begin{equation}
    \begin{aligned}
    T_R = Decoder(Concate(T_V^E,T_M)),
    \end{aligned}
\end{equation}
where $T_R \in \mathcal{R}^{L \times M \times C}$ is reconstructed part tokens.

Among the $T_R$, only the tokens masked before will be fed to the reconstruction head to predict the original masked part patches. The structure of the reconstruction head is similar to that in Point-MAE~\cite{pang2022masked}, which is a fully connected (FC) layer with a reshape operation.
\begin{equation}
    \begin{aligned}
    P^s_R = Reshape(FC(T^s_R)), \\
    P^t_R = Reshape(FC(T^t_R)), 
    \end{aligned}
\end{equation}
where $P^s_R \in \mathcal{R}^{(1-r_t)L \times r_sM \times N' \times D}$, $T^s_R \in \mathcal{R}^{(1-r_t)L \times r_sM \times C}$, $P^t_R \in \mathcal{R}^{r_tL \times M \times N' \times D}$, $T^t_R \in \mathcal{R}^{r_tL \times M \times C}$ are spatial reconstructed part patches, spatial reconstructed part tokens, temporal reconstructed part patches, temporal reconstructed part tokens, respectively.
We first reconstruct the spatial masked tokens, and then the temporal masked tokens. We adopt Chamfer Distance~\cite{fan2017point} Loss $L_{CD}$ as the reconstruction loss function:

\begin{equation}
    \begin{aligned}
    L_{CD} (P_M,P_R) = \frac{1}{\|P_M\|}\sum_{x \in P_M}\min_{y \in P_R}\|x-y\|_{2}^2 \\
    +\frac{1}{\|P_R\|}\sum_{y \in P_R}\min_{x \in P_M}\| y-x \|_{2}^2.
    \end{aligned}
\end{equation}

\subsection{Hierarchical Feature Enhanced Fine-tuning}
\label{subsec:finetine}
After the above self-learning process, STEncoder is endowed with the ability to extract the representations of humans in part-level structural semantics. When fine-tuning on multiple downstream tasks, hierarchical features can enhance the STEncoder to capture more complicated and challenging fine-grained geometric and dynamic representations.
Specifically, we integrate global-level, part-level, and point-level point cloud features to pre-trained STEncoder (See Figure.~\ref{fig:main_figure}), therefore fully leveraging prior knowledge for effective and robust human-centric representation learning.

During this stage, all part patches are visible to the STEncoder, and information of point-wise motion flow vector will be fused to that of body parts in Tokenizer. (See details in supplementary materials)
\begin{equation}
    \begin{aligned}
    T = Tokenizer(P,F),
    \end{aligned}
\end{equation}
where $T \in \mathcal{R}^{L \times M \times C}, P \in \mathcal{R}^{L \times M \times N' \times D}, F \in \mathcal{R}^{L \times M \times N' \times D'}$ are part tokens, part patches, and motion flow vectors, respectively. We will also append a global token extracted from the entire human instance to enable the interaction of features between global and part. For classification tasks like action recognition, a class token will be appended as well.
we discard the decoder and add corresponding task heads after the pre-trained tokenizer, learnable position encoding, and STEncoder for different tasks.

\section{Experiments}
\label{sec:experiments}
\begin{table*}\scriptsize 
\caption{Action Recognition in HuCenLife~\cite{xu2023human}.  $^{\dag}$ means adding global token and motion flow to these methods fair comparisons.}
\begin{tabular}{l|r|r|r|r|r|r|r|r|r|r|r|r|r}
\Xhline{1px}
                       & \multicolumn{1}{l|}{lift} & \multicolumn{1}{l|}{carry} & \multicolumn{1}{l|}{move} & \multicolumn{1}{l|}{pull\_push} & \multicolumn{1}{l|}{sco-bal} & \multicolumn{1}{l|}{hum-inter} & \multicolumn{1}{l|}{fitness} & \multicolumn{1}{l|}{entertain} & \multicolumn{1}{l|}{sports} & \multicolumn{1}{l|}{bend-over} & \multicolumn{1}{l|}{sit} & \multicolumn{1}{l|}{walk-stand} & \multicolumn{1}{l}{mAcc} \\ \hline
PointNet~\cite{qi2017pointnet}              & 45.5& 48.8& 33.3& 84& 59.4& 2.6& 65.3& 49.3& 34.8& 29.2& 54.3& 61& 47.3
\\ \hline
PointNet++~\cite{qi2017pointnet++}            & 49.5& 45.7& 35.6& 52.7& 59& 6& 28.6& 43.8& 41.2& 31.9& 38.8& 55& 40.7
\\ \hline
PointMLP~\cite{ma2022rethinking}              & 48.5& 47.7& 57.7& 80.1& 80.3& 36.1& 75.7& 60.8& 39.5& 54.9& 55.8& 59.7& 58.1
\\ \hline
PointNeXt~\cite{qian2022pointnext}             & 48.1& 56.6& 34.1& 80& 85.6& 22.6& 50& 38& 25.7& 25.5& 63.1& 70.9& 50
\\ \hline
PCT~\cite{guo2021pct} & 39.7& 54.9& 52.3& 80.2& 89.8& 9.8& 63.3& 73.6& 37.7& 62.5& 51& 75.8& 57.6
\\ \hline
HuCenLife~\cite{xu2023human}             & 45& 44.4& 52.7& 81.2& 86.7& 23.1& 81.2& 54.8& 41.7& 54.8& 53.2& 70& 57.4
\\ \hline
PointMAE$^{\dag}$~\cite{pang2022masked} & 53.4& 53.1& 47.2& 84.9& 88.8& 7.8& 71.4& 76.8& 39.2& 57.9& 41.8& 74.2& 58.0
\\ \hline
MaST-Pre$^{\dag}$~\cite{shen2023masked}  & 32.8& 39.9& 48.4& 84.5& 87.4& 31.4& 70.7& 59.1& 43.3& 51.7& 66.9& 32.5& 54.1
\\ \hline

\textbf{UniPVU-Human}& 27.1& 37.3& 57.1& 82.6& 84& 24.7& 85.4& 52.1& 53.9& 93.8& 67.3& 76.1& \textbf{61.8}\\ \Xhline{1px}
\end{tabular}
\label{experiment:hucenlife}
\end{table*}

To evaluate the effectiveness of our method, we conduct experiments on open datasets on tasks of human action recognition and human pose estimation.  
Extensive ablation studies are also conducted for the comprehensive evaluation of modules and technical designs of our method. 

\subsection{Datasets}
In this section, we first introduce our two synthetic datasets for body segmentation and motion flow estimation. Then we give details for two open datasets, which are used for evaluating our unified framework on downstream real-life human-centric tasks.

\textbf{Human Body Segmentation Synthetic Dataset. }
\label{subsubsec:HBPSegData}
To address the absence of 3D human body part segmentation datasets based on LiDAR point clouds, we leverage SMPL mesh from AMASS~\cite{mahmood2019amass} to simulate 1 million LiDAR human point cloud instances following LIP~\cite{ren2023lidar}. We automatically label the data by utilizing the SMPL mesh properties. Specifically, since ordered and regular SMPL mesh vertices provide 24 human body part labels, we can automatically assign each simulated LiDAR point the label of its nearest vertex. Due to the sparsity of point clouds, there tend to be fewer points for some body parts such as hands or feet, so we simplify the original 24 body part labels in the mesh vertices to 9, including head, left arm, right arm, upper body, lower body, left upper leg, left lower leg, right upper leg, and right lower leg.
In practical applications, occlusions and noise are inevitable. To address these challenges, we synthesize point clouds in various shapes and attach them to the appropriate positions of human point clouds to simulate common noises, such as carrying objects or using umbrellas. Subsequently, these points are labeled as ``noise". Additionally, we randomly crop the human point clouds to mimic occlusions. These operations enable our human body segmentation network to distinguish noise and adapt to occlusions, thus improving its robustness and performance.

\textbf{Human Motion Flow Synthetic Dataset. }
\label{subsubsec:HMFlowData}
Based on the LiDAR point cloud generation model~\cite{ren2023lidar,cong2021input}, we create a motion flow estimation synthetic dataset. 
We derive human motion flow by matching irregular and unordered LiDAR point clouds with regular and ordered mesh vertices, thereby establishing a correspondence between synthetic points in consecutive frames. 
We follow LIP~\cite{ren2023lidar} to create 2,378,871 frames of synthetic point clouds from SMPL mesh in AMASS~\cite{mahmood2019amass} and SURREAL~\cite{varol2017learning}, which provide diverse human motions. We match each point to the nearest mesh vertex by utilizing the k-Nearest Neighborhood (kNN) algorithm. Since the vertices of each frame have one-to-one correspondences, we can find the corresponding points in each frame based on the matched vertices. Moreover, we also use bidirectional filtering and set distance thresholds to improve the accuracy of finding corresponding points.

\textbf{Action Recognition Dataset. }
\textbf{HuCenLife}~\cite{xu2023human} is a human-centric action recognition dataset. It comprises 65,265 human instances and 12 kinds of human actions. We divide the dataset into 27142 partially overlapping clips containing 30 consecutive frames. 19594 clips are used as the train set while 7548 clips are used as the test set. It adopts the class mean accuracy (mAcc) as the evaluation metric.

\textbf{3D Pose Estimation Dataset. }
\textbf{LIPD}~\cite{ren2023lidar} is a long-range LiDAR-IMU hybrid human mocap dataset with diverse challenge motions. It comprises 15 performers with 30 types of motions, totaling 62,341 LiDAR point cloud frames, each paired with corresponding IMU measurements. Following the LIP~\cite{ren2023lidar} protocol, we divide the dataset into 39,593 frames for training and 22,748 frames for testing. We use mean per root-relative joint position error (MPJPE) in millimeters as the evaluation metric.


\begin{table}\scriptsize
\centering
\caption{3D Pose Estimation in LIP~\cite{ren2023lidar}.}
\label{experiment:pose estimation}
\begin{tabular}{l|c}
\Xhline{1px}
           & \multicolumn{1}{l}{MPJPE(mm)$\downarrow$} \\ \hline 
LiDARCap(PC)~\cite{li2022lidarcap}          & 69.4                           \\ \hline \hline
LIP(PC)~\cite{ren2023lidar} & 60.1                           \\ \hline
\textbf{UniPVU-Human(PC)}& \textbf{58.8}                          \\ \hline \hline
LIP(PC+IMU)~\cite{ren2023lidar} & 48.9                           \\ \hline
\textbf{UniPVU-Human(PC+IMU)}& \textbf{47.2}                          \\ 

\Xhline{1px}
\end{tabular}
\end{table}



\subsection{Implementation Details}
\label{subsec:pretraining}

For HuCenLife, we use point clouds with consecutive frames of $L=30$ frames as the input ($L=32$ when dealing with LIP). The dimension $D$ of each point is 3. After normalizing and sampling to $N=384$ points by Farthest Point Sampling (FPS), we apply pre-trained HBSeg and HMFlow on real data to obtain the point-wise segmentation labels $S$ and motion flow vectors $F$ with dimension $D'$ set to 3. 

For each instance, we group points with the same part segmentation labels into 9 point patches, denoted as $P$, with $M$ representing the total number of patches.
Subsequently, $N' = 48$ points are sampled using FPS in each point patch $P$. 
After being masked with temporal mask ratio $r_t=0.8$ and spatial mask ratio $r_s = 0.6$, each visible point patch has its part token embedding derived using Mini-PointNet \cite{qi2017pointnet}, with a channel dimension $C=384$.
In the self-learning stage, the features of $F$ are not used, to prevent the premature leakage of location information of masked tokens to the STEncoder. 
The spatial and temporal positional encoding are obtained by applying MLP to the average coordinate of $P$ and to the time index ranging from 0 to $L$-1, respectively. 
For STEncoder, we set the number of heads to 6. It contains 4 spatial, 4 temporal, and 4 spatial transformer layers sequentially. 
The decoder consists of 4 spatial transformer layers, and ChamferDistanceL2 is used as the loss function for mask reconstruction.
AdamW optimizer is used with an initial learning rate of 0.001 and a weight decay of 0.05. The model is trained for 300 epochs with a batch size of 512.

During fine-tuning, $P$ and corresponding motion flow vector $F$ are extracted by two tokenizers respectively and element-wise added in latent space (See details in supplementary materials), with a global token extracted from the entire human and a class token appended. Totally 11 tokens are sent to the pre-trained STEncoder. 
The pre-trained model is fine-tuned for 100 epochs using a batch size of 256 on 4 GPUs.  We use the AdamW optimizer, and the initial learning rate is set to 0.0005 with a cosine decay strategy.

\subsection{Results}
\subsubsection{Action Recognition in HuCenLife}

The results of all methods on HuCenLife are shown in Table.\ref{experiment:hucenlife}, and the evaluation metric is class mean accuracy (mAcc). For the first seven methods in the table, which are designed for processing static point clouds, we apply them on each frame of the point cloud sequence and then fuse these frame features after the encoder network by element-wise adding. For methods that also use self-learning like PointMAE~\cite{pang2022masked} and MaST-Pre~\cite{shen2023masked}, we add a global token and motion flow for fair comparisons. The experimental results demonstrate that our method significantly outperforms the methods that do not utilize self-learning. Even against general static point cloud methods with self-learning, like PointMAE, our approach shows a marked improvement due to our comprehensive temporal dynamic modeling. Compared to methods like MaST-Pre, which also models both temporal and spatial dimensions of point cloud videos along with mask prediction, our method still maintains a substantial advantage. This advantage is particularly evident in some action categories where accurate recognition hinges on extracting not just geometric features but also motion characteristics, including Fitness, Sports (encompassing activities like basketball and badminton), Bend-Over, and Walk-Stand. Experimental data shows that UniPVU-Human demonstrates superior recognition performance in these action categories. This highlights the effectiveness of utilizing human body prior knowledge and underscores the powerful capability of our model to fully leverage this prior knowledge for enhanced performance.

\subsubsection{3D Pose Estimation in LIP}
The Mean Per Root-Relative Joint Position Error (MPJPE) in millimeters is used as the evaluation metric, where a smaller value indicates better performance. Unlike action recognition, pose estimation in long point cloud videos focuses on long-term joint movement consistency. Our UniPVU-Human leverages a self-learning module for crucial human motion representation and enhances motion detail with motion flow during fine-tuning.
To ensure a fair comparison, we establish two settings as shown in Table.~\ref{experiment:pose estimation}.
\textbf{\textcolor[RGB]{230,150,0}{\textbf{1)}}  Pure PC: } involves only pure point clouds as input, where our method's MPJPE (58.8) is 1.3 lower than that of LIP (60.1).
\textbf{\textcolor[RGB]{230,150,0}{\textbf{2)}}  PC+IMU: } involves using both point cloud and IMU data as inputs. For UniPVU-Human, we replace the PointNet used for point cloud feature extraction in LIP with our model.
The experimental results indicate that our method's MPJPE (47.2) is 1.7 lower than that of LIP (48.9) of full configuration. 
Our method achieves SOTA performance under both settings, proving its superiority.


\subsection{Ablation Studies}
All ablation studies are conducted on HuCenLife~\cite{xu2023human}, for it is collected in real-life scenarios, making it more relevant to real-world applications.
The results are shown in Table.~\ref{experimant:design}.

Our UniPVU-human exploits both the structural semantics of human body and human motion to facilitate the acquisition of human-specific features by adopting human body parts as local patches for following spatio-temporal modeling with Transformers, unlike other methods which cluster the neighbor point clouds around the kernels sampled by FPS algorithm.
As shown in the first and second lines of~\ref{experimant:design}, our setting is identical to PointMAE ~\cite{pang2022masked} both with and without self-learning. The mean accuracies (mAcc) in these settings are 53.4\% and 56.1\%, respectively.
By enhancing PointMAE with hierarchical features in the fine-tuning stage, as shown in the third line of the table, the mAcc reaches 58\%, yielding a performance gain of 1.9\%. This indicates that hierarchical human-related features can also enhance performance in other models.
However, this setting still trails our best model by 3.8\%, underscoring the effectiveness of our model's approach of using body parts as semantic tokens.

Lines 5 to 7 of the table demonstrate that our model benefits from predicting masked tokens in both spatial and temporal dimensions within the self-learning module. Adding our self-learning module resulted in a substantial 6.2\% improvement in total. 

During the aforementioned self-learning module, our model has already achieved the capability to model the semantic structure of the human body at the part level. 
In the Hierarchical Feature Enhanced Fine-tuning module, we incorporate a global token and point-wise motion flow. As indicated in lines 8 to 10, the inclusion of these two designs leads to a notable performance improvement, highlighting the critical role of hierarchical human-related prior knowledge in enhancing the extraction of human geometric and motion representations. 
In conclusion, by seamlessly integrating various elements of our design, UniPVU-Human achieves exceptional performance with a final accuracy of 61.8\%. This demonstrates the effectiveness of our harmonious incorporation of components in facilitating human-centric representation learning.

\begin{table}\scriptsize
\caption{Ablation Studies of Network Design. To assess the effectiveness of designs in our UniPVU-Human, we perform ablation experiments by adding (\textcolor{green}{\checkmark}) or removing (\textcolor{red}{\XSolidBrush}) them, and then present the corresponding resulting changes in performance on HuCenLife~\cite{xu2023human}.}
\begin{tabular}{c|cc|cc|c}
\Xhline{1px}
\multirow{2}{*}{part division} & \multicolumn{2}{c|}{Self-learning Mask}  & \multicolumn{2}{c|}{Hierarchical Feature} & \multirow{2}{*}{mAcc} \\ \cline{2-5}
                               & \multicolumn{1}{c|}{spatial} & temporal & \multicolumn{1}{c|}{global token}   & motion flow  &                       \\ \hline
\textcolor{red}{\XSolidBrush}                              & \multicolumn{1}{c|} {\textcolor{red}{\XSolidBrush} }            & \textcolor{red}{\XSolidBrush}             & \multicolumn{1}{c|}{\textcolor{red}{\XSolidBrush} }              & \textcolor{red}{\XSolidBrush}            & 53.4                  \\ \hline
\textcolor{red}{\XSolidBrush}                              & \multicolumn{1}{c|} {\textcolor{green}{\checkmark} }            & \textcolor{red}{\XSolidBrush}             & \multicolumn{1}{c|}{\textcolor{red}{\XSolidBrush} }              & \textcolor{red}{\XSolidBrush}            & 56.1                  \\ \hline
\textcolor{red}{\XSolidBrush}                              & \multicolumn{1}{c|}{\textcolor{green}{\checkmark} }            & \textcolor{red}{\XSolidBrush}             & \multicolumn{1}{c|}{\textcolor{green}{\checkmark} }              & \textcolor{green}{\checkmark}             & 58                    \\ \hline
\textcolor{green}{\checkmark}                             & \multicolumn{1}{c|}{\textcolor{red}{\XSolidBrush} }            & \textcolor{red}{\XSolidBrush}             & \multicolumn{1}{c|}{\textcolor{red}{\XSolidBrush} }              & \textcolor{red}{\XSolidBrush}            & 54.1                  \\ \hline
\textcolor{green}{\checkmark}                             & \multicolumn{1}{c|}{\textcolor{red}{\XSolidBrush} }            & \textcolor{red}{\XSolidBrush}             & \multicolumn{1}{c|}{\textcolor{green}{\checkmark} }              & \textcolor{green}{\checkmark}            & 55.6                  \\ \hline
\textcolor{green}{\checkmark}                               & \multicolumn{1}{c|}{\textcolor{green}{\checkmark} }            & \textcolor{red}{\XSolidBrush}             & \multicolumn{1}{c|}{\textcolor{green}{\checkmark} }              & \textcolor{green}{\checkmark}            & 59.9                  \\ \hline
\textcolor{green}{\checkmark}                               & \multicolumn{1}{c|}{\textcolor{red}{\XSolidBrush} }            & \textcolor{green}{\checkmark}             & \multicolumn{1}{c|}{\textcolor{green}{\checkmark} }              & \textcolor{green}{\checkmark}            & 59.2                  \\ \hline
\textcolor{green}{\checkmark}                               & \multicolumn{1}{c|}{\textcolor{green}{\checkmark} }            & \textcolor{green}{\checkmark}             & \multicolumn{1}{c|}{\textcolor{red}{\XSolidBrush} }              & \textcolor{red}{\XSolidBrush}            & 58.9                  \\ \hline
\textcolor{green}{\checkmark}                               & \multicolumn{1}{c|}{\textcolor{green}{\checkmark} }            & \textcolor{green}{\checkmark}             & \multicolumn{1}{c|}{\textcolor{red}{\XSolidBrush} }              & \textcolor{green}{\checkmark}            & 59.3                  \\ \hline
\textcolor{green}{\checkmark}                              & \multicolumn{1}{c|}{\textcolor{green}{\checkmark} }            & \textcolor{green}{\checkmark}             & \multicolumn{1}{c|}{\textcolor{green}{\checkmark} }              & \textcolor{red}{\XSolidBrush}            & 61.3                  \\ \hline
\textcolor{green}{\checkmark}                              & \multicolumn{1}{c|}{\textcolor{green}{\checkmark} }            & \textcolor{green}{\checkmark}             & \multicolumn{1}{c|}{\textcolor{green}{\checkmark} }              & \textcolor{green}{\checkmark}            & \textbf{61.8}                  \\ \Xhline{1px}
\end{tabular}
\label{experimant:design}
\end{table}
\subsection{Effectiveness of Our Self-learning Mechanism in Semi-supervised Settings}
For supervised learning methods, the optimization targets of neural networks mainly come from human annotations. To endow models with strong robustness and generalizability for diverse applications, extensive data annotation is usually required. Therefore, our method introduces a self-learning mechanism, which diminishes the dependency on manual annotations, allowing our model to undergo self-learning on a vast quantity of unannotated data. Additionally, our method learns intrinsic representations directly from the data, uninhibited by the constraints and biases of task-specific, scenario-specific, or dataset-specific manual annotations. As a result, the representations acquired are not task-specific and exhibit strong generalization capabilities, which improves performance on downstream tasks. 

To validate this, we randomly sample the training set of HuCenLife by categories and assess our model's performance in fine-tuning with reduced data volumes, thereby verifying the effectiveness of self-learning. The experimental results in Table.~\ref{experimant:proportion} illustrate that when downsampling the downstream task dataset HuCenLife to 20\%, 30\%, and 50\%, our method shows the least decline in performance compared to UniPVU-Human without self-learning and MaST-Pre.


\begin{table}\footnotesize
\caption{Effectiveness of Our Self-learning Mechanism in Semi-supervised Settings on HuCenLife. * means training on HuCenLife directly without the self-learning stage. The experimental results indicate that our method demonstrates the smallest decline in performance. }
\begin{tabular}{l|llll}
\Xhline{1px}
\multirow{2}{*}{}    & \multicolumn{4}{c}{proportion of fine-tuning dataset}                                                         \\ \cline{2-5} 
                           & \multicolumn{1}{r|}{20\%} & \multicolumn{1}{r|}{30\%} & \multicolumn{1}{r|}{50\%} & \multicolumn{1}{r}{100\%} \\ \hline
MaST-Pre~\cite{shen2023masked}& \multicolumn{1}{r|}{39.8(-14.3)}     & \multicolumn{1}{r|}{42(-12.1)}     & \multicolumn{1}{r|}{48.8(-5.3)}     &                            54.1
\\ \hline
UniPVU-Human*& \multicolumn{1}{r|}{44.9(-10.9)}     & \multicolumn{1}{r|}{46.4(-9.4)}     & \multicolumn{1}{r|}{49.5(-6.3)}     &                            55.8
\\ \hline
\textbf{UniPVU-Human}& \multicolumn{1}{r|}{51(\textbf{-10.8})} & \multicolumn{1}{r|}{53.8(\textbf{-8})} & \multicolumn{1}{r|}{57.3(\textbf{-4.5})} & \multicolumn{1}{l}{61.8}  \\ \Xhline{1px}
\end{tabular}
\label{experimant:proportion}
\end{table}
\section{Conclusion}
Given the distinctive characteristics inherent to humans, including the structural semantics of the human body and the dynamics of human motions, we introduce a novel method in this paper to delve into the intrinsic features present within the data itself to facilitate a more comprehensive understanding of human-centric point cloud videos. To our knowledge, our method is the first work to provide a unified framework designed specifically for tackling human-centric tasks. Extensive experiments on various tasks have demonstrated the state-of-the-art performance of our method.  
{
    \small
    \bibliographystyle{ieeenat_fullname}
    \bibliography{arxiv}

\begin{thebibliography}{50}
\providecommand{\natexlab}[1]{#1}
\providecommand{\url}[1]{\texttt{#1}}
\expandafter\ifx\csname urlstyle\endcsname\relax
  \providecommand{\doi}[1]{doi: #1}\else
  \providecommand{\doi}{doi: \begingroup \urlstyle{rm}\Url}\fi

\bibitem[Chen et~al.(2023)Chen, Wang, Yang, Yu, Yuan, and Yue]{chen2023pointgpt}
Guangyan Chen, Meiling Wang, Yi Yang, Kai Yu, Li Yuan, and Yufeng Yue.
\newblock Pointgpt: Auto-regressively generative pre-training from point clouds.
\newblock \emph{arXiv preprint arXiv:2305.11487}, 2023.

\bibitem[Choy et~al.(2019)Choy, Gwak, and Savarese]{choy20194d}
Christopher Choy, JunYoung Gwak, and Silvio Savarese.
\newblock 4d spatio-temporal convnets: Minkowski convolutional neural networks.
\newblock In \emph{Proceedings of the IEEE/CVF conference on computer vision and pattern recognition}, pages 3075--3084, 2019.

\bibitem[Cong et~al.(2021)Cong, Zhu, and Ma]{cong2021input}
Peishan Cong, Xinge Zhu, and Yuexin Ma.
\newblock Input-output balanced framework for long-tailed lidar semantic segmentation.
\newblock In \emph{2021 IEEE International Conference on Multimedia and Expo (ICME)}, pages 1--6. IEEE, 2021.

\bibitem[Cong et~al.(2023)Cong, Xu, Ren, Zhang, Xu, Wang, Yu, and Ma]{cong2023weakly}
Peishan Cong, Yiteng Xu, Yiming Ren, Juze Zhang, Lan Xu, Jingya Wang, Jingyi Yu, and Yuexin Ma.
\newblock Weakly supervised 3d multi-person pose estimation for large-scale scenes based on monocular camera and single lidar.
\newblock In \emph{Proceedings of the AAAI Conference on Artificial Intelligence}, pages 461--469, 2023.

\bibitem[Dai et~al.(2023)Dai, Lin, Lin, Wen, Xu, Yi, Shen, Ma, and Wang]{dai2023sloper4d}
Yudi Dai, YiTai Lin, XiPing Lin, Chenglu Wen, Lan Xu, Hongwei Yi, Siqi Shen, Yuexin Ma, and Cheng Wang.
\newblock Sloper4d: A scene-aware dataset for global 4d human pose estimation in urban environments.
\newblock In \emph{Proceedings of the IEEE/CVF Conference on Computer Vision and Pattern Recognition}, pages 682--692, 2023.

\bibitem[Fan et~al.(2017)Fan, Su, and Guibas]{fan2017point}
Haoqiang Fan, Hao Su, and Leonidas~J Guibas.
\newblock A point set generation network for 3d object reconstruction from a single image.
\newblock In \emph{Proceedings of the IEEE conference on computer vision and pattern recognition}, pages 605--613, 2017.

\bibitem[Fan et~al.(2021{\natexlab{a}})Fan, Yang, and Kankanhalli]{fan2021point}
Hehe Fan, Yi Yang, and Mohan Kankanhalli.
\newblock Point 4d transformer networks for spatio-temporal modeling in point cloud videos.
\newblock In \emph{Proceedings of the IEEE/CVF conference on computer vision and pattern recognition}, pages 14204--14213, 2021{\natexlab{a}}.

\bibitem[Fan et~al.(2021{\natexlab{b}})Fan, Yu, Yang, and Kankanhalli]{fan2021deep}
Hehe Fan, Xin Yu, Yi Yang, and Mohan Kankanhalli.
\newblock Deep hierarchical representation of point cloud videos via spatio-temporal decomposition.
\newblock \emph{IEEE Transactions on Pattern Analysis and Machine Intelligence}, 44\penalty0 (12):\penalty0 9918--9930, 2021{\natexlab{b}}.

\bibitem[Fan et~al.(2022{\natexlab{a}})Fan, Yang, and Kankanhalli]{fan2022point}
Hehe Fan, Yi Yang, and Mohan Kankanhalli.
\newblock Point spatio-temporal transformer networks for point cloud video modeling.
\newblock \emph{IEEE Transactions on Pattern Analysis and Machine Intelligence}, 45\penalty0 (2):\penalty0 2181--2192, 2022{\natexlab{a}}.

\bibitem[Fan et~al.(2022{\natexlab{b}})Fan, Yu, Ding, Yang, and Kankanhalli]{fan2022pstnet}
Hehe Fan, Xin Yu, Yuhang Ding, Yi Yang, and Mohan Kankanhalli.
\newblock Pstnet: Point spatio-temporal convolution on point cloud sequences.
\newblock \emph{arXiv preprint arXiv:2205.13713}, 2022{\natexlab{b}}.

\bibitem[Gu et~al.(2019)Gu, Wang, Wu, Lee, and Wang]{gu2019hplflownet}
Xiuye Gu, Yijie Wang, Chongruo Wu, Yong~Jae Lee, and Panqu Wang.
\newblock Hplflownet: Hierarchical permutohedral lattice flownet for scene flow estimation on large-scale point clouds.
\newblock In \emph{Proceedings of the IEEE/CVF conference on computer vision and pattern recognition}, pages 3254--3263, 2019.

\bibitem[Guo et~al.(2021)Guo, Cai, Liu, Mu, Martin, and Hu]{guo2021pct}
Meng-Hao Guo, Jun-Xiong Cai, Zheng-Ning Liu, Tai-Jiang Mu, Ralph~R Martin, and Shi-Min Hu.
\newblock Pct: Point cloud transformer.
\newblock \emph{Computational Visual Media}, 7:\penalty0 187--199, 2021.

\bibitem[Han et~al.(2022)Han, Cong, Xu, Wang, Yu, and Ma]{han2022licamgait}
Xiao Han, Peishan Cong, Lan Xu, Jingya Wang, Jingyi Yu, and Yuexin Ma.
\newblock Licamgait: Gait recognition in the wild by using lidar and camera multi-modal visual sensors.
\newblock \emph{arXiv preprint arXiv:2211.12371}, 2022.

\bibitem[He et~al.(2016)He, Zhang, Ren, and Sun]{he2016deep}
Kaiming He, Xiangyu Zhang, Shaoqing Ren, and Jian Sun.
\newblock Deep residual learning for image recognition.
\newblock In \emph{Proceedings of the IEEE conference on computer vision and pattern recognition}, pages 770--778, 2016.

\bibitem[He et~al.(2022)He, Chen, Xie, Li, Doll{\'a}r, and Girshick]{he2022masked}
Kaiming He, Xinlei Chen, Saining Xie, Yanghao Li, Piotr Doll{\'a}r, and Ross Girshick.
\newblock Masked autoencoders are scalable vision learners.
\newblock In \emph{Proceedings of the IEEE/CVF conference on computer vision and pattern recognition}, pages 16000--16009, 2022.

\bibitem[Li et~al.(2022)Li, Zhang, Wang, Shen, Wen, Ma, Xu, Yu, and Wang]{li2022lidarcap}
Jialian Li, Jingyi Zhang, Zhiyong Wang, Siqi Shen, Chenglu Wen, Yuexin Ma, Lan Xu, Jingyi Yu, and Cheng Wang.
\newblock Lidarcap: Long-range marker-less 3d human motion capture with lidar point clouds.
\newblock In \emph{Proceedings of the IEEE/CVF Conference on Computer Vision and Pattern Recognition}, pages 20502--20512, 2022.

\bibitem[Li et~al.(2023)Li, Xiang, Chen, Zhang, and Yang]{li2023deep}
Zhiqi Li, Nan Xiang, Honghua Chen, Jianjun Zhang, and Xiaosong Yang.
\newblock Deep learning for scene flow estimation on point clouds: A survey and prospective trends.
\newblock In \emph{Computer Graphics Forum}. Wiley Online Library, 2023.

\bibitem[Liu et~al.(2019{\natexlab{a}})Liu, Qi, and Guibas]{liu2019flownet3d}
Xingyu Liu, Charles~R Qi, and Leonidas~J Guibas.
\newblock Flownet3d: Learning scene flow in 3d point clouds.
\newblock In \emph{Proceedings of the IEEE/CVF conference on computer vision and pattern recognition}, pages 529--537, 2019{\natexlab{a}}.

\bibitem[Liu et~al.(2019{\natexlab{b}})Liu, Yan, and Bohg]{liu2019meteornet}
Xingyu Liu, Mengyuan Yan, and Jeannette Bohg.
\newblock Meteornet: Deep learning on dynamic 3d point cloud sequences.
\newblock In \emph{Proceedings of the IEEE/CVF International Conference on Computer Vision}, pages 9246--9255, 2019{\natexlab{b}}.

\bibitem[Liu et~al.(2023)Liu, Tian, Lv, Li, and Wang]{liu2023point}
Yahui Liu, Bin Tian, Yisheng Lv, Lingxi Li, and Fei-Yue Wang.
\newblock Point cloud classification using content-based transformer via clustering in feature space.
\newblock \emph{IEEE/CAA Journal of Automatica Sinica}, 2023.

\bibitem[Loper et~al.(2023)Loper, Mahmood, Romero, Pons-Moll, and Black]{loper2023smpl}
Matthew Loper, Naureen Mahmood, Javier Romero, Gerard Pons-Moll, and Michael~J Black.
\newblock Smpl: A skinned multi-person linear model.
\newblock In \emph{Seminal Graphics Papers: Pushing the Boundaries, Volume 2}, pages 851--866. 2023.

\bibitem[Ma et~al.(2022)Ma, Qin, You, Ran, and Fu]{ma2022rethinking}
Xu Ma, Can Qin, Haoxuan You, Haoxi Ran, and Yun Fu.
\newblock Rethinking network design and local geometry in point cloud: A simple residual mlp framework.
\newblock \emph{arXiv preprint arXiv:2202.07123}, 2022.

\bibitem[Mahmood et~al.(2019)Mahmood, Ghorbani, Troje, Pons-Moll, and Black]{mahmood2019amass}
Naureen Mahmood, Nima Ghorbani, Nikolaus~F Troje, Gerard Pons-Moll, and Michael~J Black.
\newblock Amass: Archive of motion capture as surface shapes.
\newblock In \emph{Proceedings of the IEEE/CVF international conference on computer vision}, pages 5442--5451, 2019.

\bibitem[Pang et~al.(2022)Pang, Wang, Tay, Liu, Tian, and Yuan]{pang2022masked}
Yatian Pang, Wenxiao Wang, Francis~EH Tay, Wei Liu, Yonghong Tian, and Li Yuan.
\newblock Masked autoencoders for point cloud self-supervised learning.
\newblock In \emph{European conference on computer vision}, pages 604--621. Springer, 2022.

\bibitem[Puy et~al.(2020)Puy, Boulch, and Marlet]{puy2020flot}
Gilles Puy, Alexandre Boulch, and Renaud Marlet.
\newblock Flot: Scene flow on point clouds guided by optimal transport.
\newblock In \emph{European conference on computer vision}, pages 527--544. Springer, 2020.

\bibitem[Qi et~al.(2017{\natexlab{a}})Qi, Su, Mo, and Guibas]{qi2017pointnet}
Charles~R Qi, Hao Su, Kaichun Mo, and Leonidas~J Guibas.
\newblock Pointnet: Deep learning on point sets for 3d classification and segmentation.
\newblock In \emph{Proceedings of the IEEE conference on computer vision and pattern recognition}, pages 652--660, 2017{\natexlab{a}}.

\bibitem[Qi et~al.(2017{\natexlab{b}})Qi, Yi, Su, and Guibas]{qi2017pointnet++}
Charles~Ruizhongtai Qi, Li Yi, Hao Su, and Leonidas~J Guibas.
\newblock Pointnet++: Deep hierarchical feature learning on point sets in a metric space.
\newblock \emph{Advances in neural information processing systems}, 30, 2017{\natexlab{b}}.

\bibitem[Qian et~al.(2022)Qian, Li, Peng, Mai, Hammoud, Elhoseiny, and Ghanem]{qian2022pointnext}
Guocheng Qian, Yuchen Li, Houwen Peng, Jinjie Mai, Hasan Hammoud, Mohamed Elhoseiny, and Bernard Ghanem.
\newblock Pointnext: Revisiting pointnet++ with improved training and scaling strategies.
\newblock \emph{Advances in Neural Information Processing Systems}, 35:\penalty0 23192--23204, 2022.

\bibitem[Ren et~al.(2023)Ren, Zhao, He, Cong, Liang, Yu, Xu, and Ma]{ren2023lidar}
Yiming Ren, Chengfeng Zhao, Yannan He, Peishan Cong, Han Liang, Jingyi Yu, Lan Xu, and Yuexin Ma.
\newblock Lidar-aid inertial poser: Large-scale human motion capture by sparse inertial and lidar sensors.
\newblock \emph{IEEE Transactions on Visualization and Computer Graphics}, 29\penalty0 (5):\penalty0 2337--2347, 2023.

\bibitem[Shen et~al.(2023{\natexlab{a}})Shen, Sheng, Fan, Wang, Guo, Liu, Wen, and Zhou]{shen2023masked}
Zhiqiang Shen, Xiaoxiao Sheng, Hehe Fan, Longguang Wang, Yulan Guo, Qiong Liu, Hao Wen, and Xi Zhou.
\newblock Masked spatio-temporal structure prediction for self-supervised learning on point cloud videos.
\newblock In \emph{Proceedings of the IEEE/CVF International Conference on Computer Vision}, pages 16580--16589, 2023{\natexlab{a}}.

\bibitem[Shen et~al.(2023{\natexlab{b}})Shen, Sheng, Wang, Guo, Liu, and Zhou]{shen2023pointcmp}
Zhiqiang Shen, Xiaoxiao Sheng, Longguang Wang, Yulan Guo, Qiong Liu, and Xi Zhou.
\newblock Pointcmp: Contrastive mask prediction for self-supervised learning on point cloud videos.
\newblock In \emph{Proceedings of the IEEE/CVF Conference on Computer Vision and Pattern Recognition}, pages 1212--1222, 2023{\natexlab{b}}.

\bibitem[Sheng et~al.(2023{\natexlab{a}})Sheng, Shen, and Xiao]{sheng2023contrastive}
Xiaoxiao Sheng, Zhiqiang Shen, and Gang Xiao.
\newblock Contrastive predictive autoencoders for dynamic point cloud self-supervised learning.
\newblock \emph{arXiv preprint arXiv:2305.12959}, 2023{\natexlab{a}}.

\bibitem[Sheng et~al.(2023{\natexlab{b}})Sheng, Shen, Xiao, Wang, Guo, and Fan]{sheng2023point}
Xiaoxiao Sheng, Zhiqiang Shen, Gang Xiao, Longguang Wang, Yulan Guo, and Hehe Fan.
\newblock Point contrastive prediction with semantic clustering for self-supervised learning on point cloud videos.
\newblock In \emph{Proceedings of the IEEE/CVF International Conference on Computer Vision}, pages 16515--16524, 2023{\natexlab{b}}.

\bibitem[Varol et~al.(2017)Varol, Romero, Martin, Mahmood, Black, Laptev, and Schmid]{varol2017learning}
Gul Varol, Javier Romero, Xavier Martin, Naureen Mahmood, Michael~J Black, Ivan Laptev, and Cordelia Schmid.
\newblock Learning from synthetic humans.
\newblock In \emph{Proceedings of the IEEE conference on computer vision and pattern recognition}, pages 109--117, 2017.

\bibitem[Vaswani et~al.(2017)Vaswani, Shazeer, Parmar, Uszkoreit, Jones, Gomez, Kaiser, and Polosukhin]{vaswani2017attention}
Ashish Vaswani, Noam Shazeer, Niki Parmar, Jakob Uszkoreit, Llion Jones, Aidan~N Gomez, {\L}ukasz Kaiser, and Illia Polosukhin.
\newblock Attention is all you need.
\newblock \emph{Advances in neural information processing systems}, 30, 2017.

\bibitem[Wang et~al.(2021)Wang, Yang, Rong, Feng, and Tian]{wang2021self}
Haiyan Wang, Liang Yang, Xuejian Rong, Jinglun Feng, and Yingli Tian.
\newblock Self-supervised 4d spatio-temporal feature learning via order prediction of sequential point cloud clips.
\newblock In \emph{Proceedings of the IEEE/CVF Winter Conference on Applications of Computer Vision}, pages 3762--3771, 2021.

\bibitem[Wen et~al.(2022)Wen, Liu, Huang, Duan, and Yi]{wen2022point}
Hao Wen, Yunze Liu, Jingwei Huang, Bo Duan, and Li Yi.
\newblock Point primitive transformer for long-term 4d point cloud video understanding.
\newblock In \emph{European Conference on Computer Vision}, pages 19--35. Springer, 2022.

\bibitem[Wu et~al.(2019)Wu, Wang, Li, Liu, and Fuxin]{wu2019pointpwc}
Wenxuan Wu, Zhiyuan Wang, Zhuwen Li, Wei Liu, and Li Fuxin.
\newblock Pointpwc-net: A coarse-to-fine network for supervised and self-supervised scene flow estimation on 3d point clouds.
\newblock \emph{arXiv preprint arXiv:1911.12408}, 2019.

\bibitem[Xie et~al.(2020)Xie, Gu, Guo, Qi, Guibas, and Litany]{xie2020pointcontrast}
Saining Xie, Jiatao Gu, Demi Guo, Charles~R Qi, Leonidas Guibas, and Or Litany.
\newblock Pointcontrast: Unsupervised pre-training for 3d point cloud understanding.
\newblock In \emph{Computer Vision--ECCV 2020: 16th European Conference, Glasgow, UK, August 23--28, 2020, Proceedings, Part III 16}, pages 574--591. Springer, 2020.

\bibitem[Xu et~al.(2023)Xu, Cong, Yao, Chen, Hou, Zhu, He, Yu, and Ma]{xu2023human}
Yiteng Xu, Peishan Cong, Yichen Yao, Runnan Chen, Yuenan Hou, Xinge Zhu, Xuming He, Jingyi Yu, and Yuexin Ma.
\newblock Human-centric scene understanding for 3d large-scale scenarios.
\newblock In \emph{Proceedings of the IEEE/CVF International Conference on Computer Vision}, pages 20349--20359, 2023.

\bibitem[Yan et~al.(2023)Yan, Wang, Dai, Shen, Wen, Xu, Ma, and Wang]{yan2023cimi4d}
Ming Yan, Xin Wang, Yudi Dai, Siqi Shen, Chenglu Wen, Lan Xu, Yuexin Ma, and Cheng Wang.
\newblock Cimi4d: A large multimodal climbing motion dataset under human-scene interactions.
\newblock In \emph{Proceedings of the IEEE/CVF Conference on Computer Vision and Pattern Recognition}, pages 12977--12988, 2023.

\bibitem[Yin et~al.(2021)Yin, Zhou, and Krahenbuhl]{centerpoint}
Tianwei Yin, Xingyi Zhou, and Philipp Krahenbuhl.
\newblock Center-based 3d object detection and tracking.
\newblock In \emph{CVPR}, pages 11784--11793, 2021.

\bibitem[Yu et~al.(2022)Yu, Tang, Rao, Huang, Zhou, and Lu]{yu2022point}
Xumin Yu, Lulu Tang, Yongming Rao, Tiejun Huang, Jie Zhou, and Jiwen Lu.
\newblock Point-bert: Pre-training 3d point cloud transformers with masked point modeling.
\newblock In \emph{Proceedings of the IEEE/CVF Conference on Computer Vision and Pattern Recognition}, pages 19313--19322, 2022.

\bibitem[Zeng et~al.(2023)Zeng, Wang, Nguyen, and Yue]{zeng2023self}
Changyu Zeng, Wei Wang, Anh Nguyen, and Yutao Yue.
\newblock Self-supervised learning for point cloud data: A survey.
\newblock \emph{Expert Systems with Applications}, page 121354, 2023.

\bibitem[Zhang et~al.(2022)Zhang, Guo, Gao, Fang, Zhao, Wang, Qiao, and Li]{zhang2022point}
Renrui Zhang, Ziyu Guo, Peng Gao, Rongyao Fang, Bin Zhao, Dong Wang, Yu Qiao, and Hongsheng Li.
\newblock Point-m2ae: multi-scale masked autoencoders for hierarchical point cloud pre-training.
\newblock \emph{Advances in neural information processing systems}, 35:\penalty0 27061--27074, 2022.

\bibitem[Zhang et~al.(2023)Zhang, Dong, Liu, and Yi]{zhang2023complete}
Zhuoyang Zhang, Yuhao Dong, Yunze Liu, and Li Yi.
\newblock Complete-to-partial 4d distillation for self-supervised point cloud sequence representation learning.
\newblock In \emph{Proceedings of the IEEE/CVF Conference on Computer Vision and Pattern Recognition}, pages 17661--17670, 2023.

\bibitem[Zhao et~al.(2021)Zhao, Jiang, Jia, Torr, and Koltun]{zhao2021point}
Hengshuang Zhao, Li Jiang, Jiaya Jia, Philip~HS Torr, and Vladlen Koltun.
\newblock Point transformer.
\newblock In \emph{Proceedings of the IEEE/CVF international conference on computer vision}, pages 16259--16268, 2021.

\bibitem[Zhou and Tuzel(2018)]{voxelnet}
Yin Zhou and Oncel Tuzel.
\newblock Voxelnet: End-to-end learning for point cloud based 3d object detection.
\newblock In \emph{CVPR}, pages 4490--4499. IEEE Computer Society, 2018.

\bibitem[Zhou et~al.(2022)Zhou, Zhao, Wang, Wang, and Foroosh]{centerformer}
Zixiang Zhou, Xiangchen Zhao, Yu Wang, Panqu Wang, and Hassan Foroosh.
\newblock Centerformer: Center-based transformer for 3d object detection.
\newblock In \emph{ECCV}, pages 496--513. Springer, 2022.

\bibitem[Zhu et~al.(2021)Zhu, Zhou, Wang, Hong, Li, Ma, Li, Yang, and Lin]{zhu2021cylindrical}
Xinge Zhu, Hui Zhou, Tai Wang, Fangzhou Hong, Wei Li, Yuexin Ma, Hongsheng Li, Ruigang Yang, and Dahua Lin.
\newblock Cylindrical and asymmetrical 3d convolution networks for lidar-based perception.
\newblock \emph{IEEE Transactions on Pattern Analysis and Machine Intelligence}, 44\penalty0 (10):\penalty0 6807--6822, 2021.

\end{thebibliography}
}
\clearpage
\setcounter{page}{1}
\begin{appendix}

\section{Overview}
In the supplementary materials, we provide detailed information on the generation of the two datasets used for training HMFlow and HBSeg, along with corresponding benchmarks and evaluation metrics to facilitate their use and assessment in future research. Additionally, we include the implementation details of the Tokenizer in both Self-learning and Fine-tuning stages. For more comprehensive and extensive comparisons, we have expanded our comparison experiments on HuCenLife~\cite{xu2023human} to include methods tailored for modeling dynamic point cloud videos, to demonstrate the superiority of our method in capturing human motion representations. Finally, we also provide additional details regarding the size of the UniPVU-Human model and comparisons with others.
\section{Human Motion Flow (HMFlow)}
\label{sec:hmflow}
\subsection{Implementation Details}
\begin{figure}[ht]
    \centering
    \includegraphics[width=1\columnwidth]{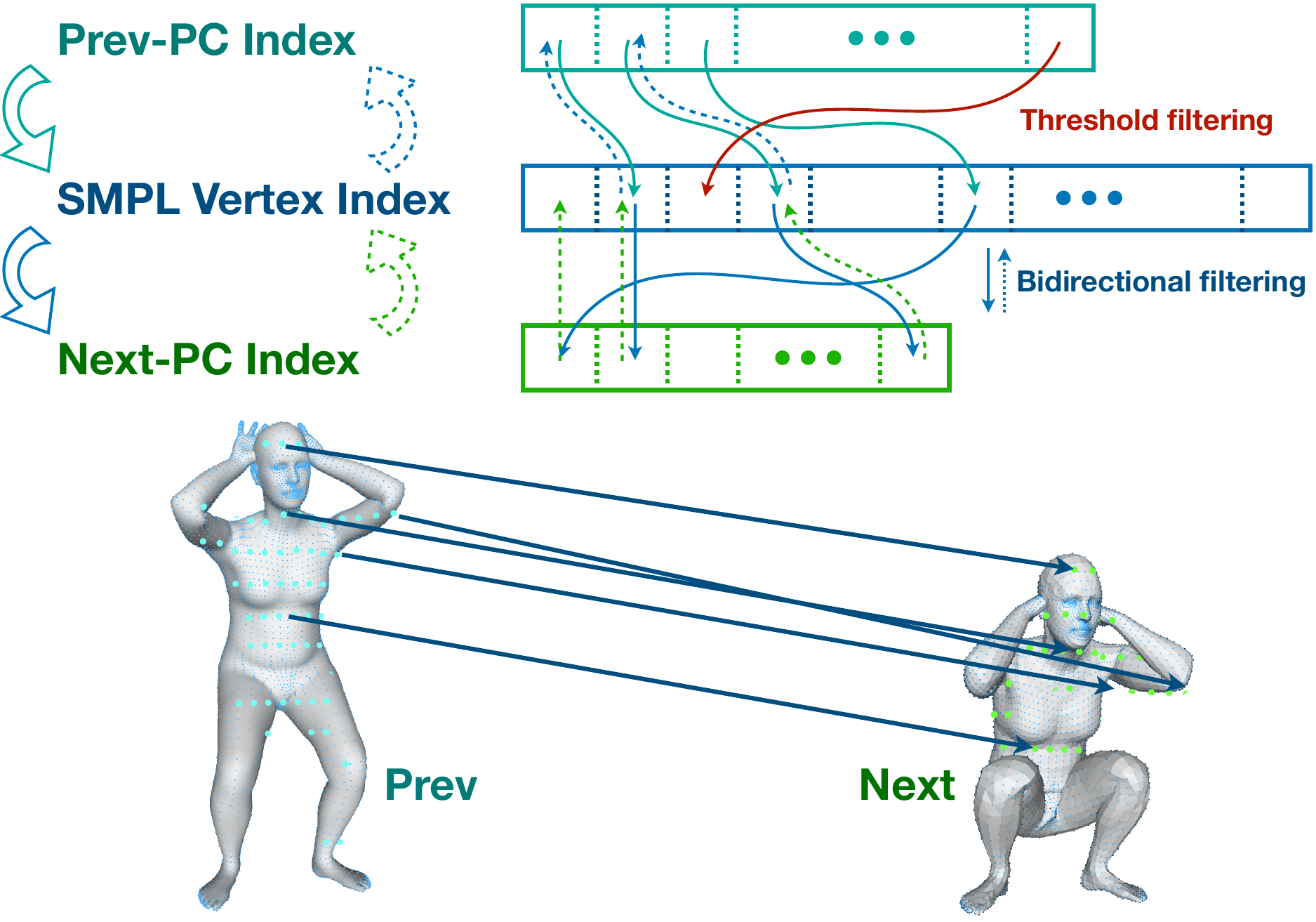}
    \caption{The pipeline of generating the flow from the previous point cloud to the next point cloud. We associate each synthetic LiDAR point to its nearest SMPL vertex, to establish the correspondence between synthetic LiDAR points across different frames by using SMPL vertices indices as medium, so that we can obtain point-wise motion flow.}
    \label{fig:flow}
    \vspace{-1ex}
\end{figure}
Due to rotation or occlusion, point clouds may flow in and out between consecutive frames, resulting in a lack of temporal correspondence. But for the SMPL~\cite{loper2023smpl} mesh, each mesh vertex can be matched between consecutive frames using vertex index. Therefore, we make a large-scale synthetic dataset, LiDARFlow-Human, by scanning the SMPL mesh surfaces of consecutive frames using a simulated LiDAR to generate simulated LiDAR point clouds (As shown in Figure.~\ref{fig:flow}). Each simulated LiDAR point is matched with its nearest SMPL vertex. Consequently, we use SMPL vertices as a medium to match simulated LiDAR points between frames. By subtracting coordinates, we obtain the point-wise motion flow. Moreover, we set a threshold to filter the distance and build the bidirectional connections to ensure the accuracy of the matching. Specifically, when the nearest distance from vertex to point is smaller than the defined threshold $D$, we think them has the unidirectional connection and we select the unidirectional connection $C_{p2n}$ from previous point cloud to next point clouds, meanwhile we select the unidirectional connection $C_{n2p}$ from next point clouds to previous point clouds. The bidirectional filter are used to delete the unidirectional connection without coincidence, $$Flow_{p2n} = C_{p2n}\cap{C_{n2p}}.$$



\subsection{Dataset and Evaluation Metrics}
We will contribute LiDARFlow-Human, used for training Human Motion Flow Estimator (HMFlow), to the community with corresponding benchmarks. As shown in Table.~\ref{tab:lidarflow_human}, we adopt the evaluation metrics used in ~\cite{puy2020flot,gu2019hplflownet,liu2019flownet3d,wu2019pointpwc}:
\begin{itemize}
\item EPE: End Point Error (meters). $$EPE = \frac{\sum_{i=1}^N \left\| (\vec{(f_{predict})_i}-\vec{(f_{gt})_i}) \right\|_2}{N},$$ where $\vec{(f_{predict})_i}$ and $\vec{(f_{gt})_i}$ are point-wise predicted motion flow and ground truth motion flow, respectively.

\item acc\_strict: percentage of points such that $EPE_i<0.05$ or $EPE_i/ \|\vec{f_i}\|_2<0.05$.
\item acc\_relax: percentage of points such that $EPE_i<0.1$ or $EPE_i/ \|\vec{f_i}\|_2<0.1$.
\item outlier: percentage of points such that $EPE_i>0.3$ or $EPE_i/ \|\vec{f_i}\|_2>0.1$.
\end{itemize}

\section{Human Body Segmentation (HBSeg)}

\begin{figure}[t]
    \centering
    \includegraphics[width=1\columnwidth]{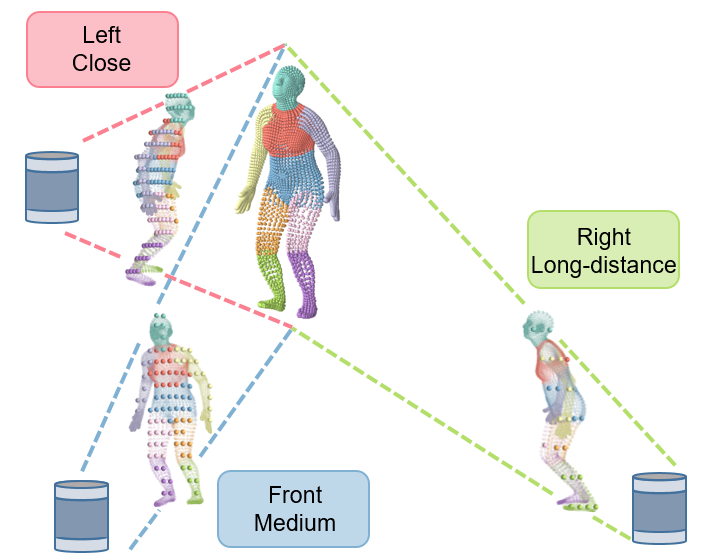}
    \caption{We create a synthetic dataset of 1 million LiDAR human point cloud instances, using the AMASS dataset for 3D human meshes and simulating LiDAR scans from various perspectives and distances for enhancing the diversity of the samples, so as to better simulate the distribution of real-world data. } 
    \label{fig:HBSeg}
    \vspace{-1ex}
\end{figure}

\begin{table}[t]
\caption{Human Motion Flow (HMFlow) Result on LiDARFlow-Human.}
\label{tab:lidarflow_human}
\scalebox{0.95}{
\begin{tabular}{c|c|c|c|c}
\Xhline{1px}
     & EPE$\downarrow$    & acc\_strict$\uparrow$ & acc\_relax$\uparrow$ & outlier$\downarrow$ \\ \hline
FLOT~\cite{puy2020flot} & 0.14 & 83.67             & 95.77           & 0.78  \\ \Xhline{1px}
\end{tabular}
}
\end{table}

\begin{table*}[]\footnotesize
\centering
\caption{Human Body Segmentation (HBSeg) Results on LiDARPart-Human.}
\label{tab:lidarpart_human}
\begin{tabular}{c|c|c|c|c|c|c|c|c|c|c}
\Xhline{1px}
           & head & left-arm & right-arm & up-body & low-body & upleft-leg & upright-leg & lowleft-leg & lowright-leg & mIoU$\uparrow$ \\ \hline
PointNet~\cite{qi2017pointnet}   & 88.2 & 51.2     & 46.6      & 52.1    & 62.6     & 45.8       & 36.2        & 67.4        & 60.2         & 56.7 \\ \hline
PointNet++~\cite{qi2017pointnet++} & 88.6 & 69.5     & 69.9      & 65.4    & 82.2     & 82.7       & 82.5        & 89.1        & 89.4         & 79.9 \\ \hline
PointMLP~\cite{ma2022rethinking}   & 92.0 & 76.1     & 75.2      & 76.7    & 88.0     & 86.3       & 85.8        & 92.8        & 92.3         & 85.0 \\ \hline
PointNeXt~\cite{qian2022pointnext}  & 95.1 & 82.7     & 81.9      & 83.1    & 91.9     & 91.2       & 90.8        & 96.1        & 96.0         & \textbf{89.9} \\ \Xhline{1px}
\end{tabular}
\end{table*}

\begin{table*}[]\scriptsize
\centering
\caption{Supplementary Comparison Experiments on HuCenLife~\cite{xu2023human}. "DM" stands for "Dynamic Method," indicating whether it is a method used for modeling dynamic point cloud videos. For the static methods, which are designed for processing static point clouds, we apply them on each frame of the point cloud sequence and then fuse these frame features after the encoder network by element-wise adding. "SL" stands for "Self-learning", signifying whether the method employs a self-learning mechanism.}
\label{tab:sup_comp_exp}.
\scalebox{0.91}{
\begin{tabular}{l|l|l|c|c|c|c|c|c|c|c|c|c|c|c|c}
\Xhline{1px}
                                                          & DM                                                 & SL                                                 & lift & carry & move & pull\_push & sco-bal & hum-inter & fitness & entertain & sports & bend-over & sit  & walk-stand & mAcc \\ \hline
PointNet~\cite{qi2017pointnet}      & \textcolor{red}{\XSolidBrush} & \textcolor{red}{\XSolidBrush} & 45.5 & 48.8  & 33.3 & 84         & 59.4    & 2.6       & 65.3    & 49.3      & 34.8   & 29.2      & 54.3 & 61         & 47.3 \\ \hline
PointNet++~\cite{qi2017pointnet++}  & \textcolor{red}{\XSolidBrush} & \textcolor{red}{\XSolidBrush} & 49.5 & 45.7  & 35.6 & 52.7       & 59      & 6         & 28.6    & 43.8      & 41.2   & 31.9      & 38.8 & 55         & 40.7 \\ \hline
PointMLP~\cite{ma2022rethinking}    & \textcolor{red}{\XSolidBrush} & \textcolor{red}{\XSolidBrush} & 48.5 & 47.7  & 57.7 & 80.1       & 80.3    & 36.1      & 75.7    & 60.8      & 39.5   & 54.9      & 55.8 & 59.7       & 58.1 \\ \hline
PointNeXt~\cite{qian2022pointnext}  & \textcolor{red}{\XSolidBrush} & \textcolor{red}{\XSolidBrush} & 48.1 & 56.6  & 34.1 & 80         & 85.6    & 22.6      & 50      & 38        & 25.7   & 25.5      & 63.1 & 70.9       & 50   \\ \hline
PCT~\cite{guo2021pct}               & \textcolor{red}{\XSolidBrush} & \textcolor{red}{\XSolidBrush} & 39.7 & 54.9  & 52.3 & 80.2       & 89.8    & 9.8       & 63.3    & 73.6      & 37.7   & 62.5      & 51   & 75.8       & 57.6 \\ \hline
HuCenLife~\cite{xu2023human}        & \textcolor{red}{\XSolidBrush} & \textcolor{red}{\XSolidBrush} & 45   & 44.4  & 52.7 & 81.2       & 86.7    & 23.1      & 81.2    & 54.8      & 41.7   & 54.8      & 53.2 & 70         & 57.4 \\ \hline
PSTNet~\cite{fan2022pstnet}         & \textcolor{green}{\checkmark} & \textcolor{red}{\XSolidBrush} &30.2	&22.6&	61.4&	64.7&	74.6	&21.6&	20.8&	82.4&	39.7	&51.1&	36	&15.4&	43.4 \\ \hline
PSTNet++~\cite{fan2021deep}         & \textcolor{green}{\checkmark} & \textcolor{red}{\XSolidBrush} &31.8&	35.4	&19.4&	77.4&	52.1&	44.8&	65.3&	52.8	&51.6	&43.8&	63&	65.3&	50.2  \\ \hline
P4Transformer~\cite{fan2021point}   & \textcolor{green}{\checkmark} & \textcolor{red}{\XSolidBrush} & 52.6 & 44.1  & 20.6 & 83.8       & 67.5    & 28.1      & 35.4    & 68.7      & 50.6   & 38.8      & 62.6 & 63.8       & 51.4 \\ \hline
PST-Transformer~\cite{fan2022point} & \textcolor{green}{\checkmark} & \textcolor{red}{\XSolidBrush} & 54.2 & 40.3  & 23.4 & 82.6       & 78.5    & 21.8      & 25      & 51.9      & 37.7   & 68.1      & 79   & 74.5       & 53.1 \\ \hline
PPTr~\cite{wen2022point}            & \textcolor{green}{\checkmark} & \textcolor{red}{\XSolidBrush} & 48.2 & 46    & 18   & 79.1       & 71.5    & 20        & 44.7    & 63.7      & 52.4   & 35.6      & 65.4 & 70         & 51.2 \\ \hline
PointMAE~\cite{pang2022masked}      & \textcolor{red}{\XSolidBrush} & \textcolor{green}{\checkmark} & 53.4 & 53.1  & 47.2 & 84.9       & 88.8    & 7.8       & 71.4    & 76.8      & 39.2   & 57.9      & 41.8 & 74.2       & 58   \\ \hline
MaST-Pre~\cite{shen2023masked}      & \textcolor{green}{\checkmark} & \textcolor{green}{\checkmark} & 32.8 & 39.9  & 48.4 & 84.5       & 87.4    & 31.4      & 70.7    & 59.1      & 43.3   & 51.7      & 66.9 & 32.5       & 54.1 \\ \hline
PointCMP~\cite{shen2023pointcmp}    & \textcolor{green}{\checkmark} & \textcolor{green}{\checkmark} & 25.6	&8.3&	56.2	&78.8	&71.9&	7.8&	65.3&	58.6	&52.9&	55.1&	72.9	&19.5&	47.7    \\ \hline
UniPVU-Human                                              & \textcolor{green}{\checkmark} & \textcolor{green}{\checkmark} & 27.1 & 37.3  & 57.1 & 82.6       & 84      & 24.7      & \textbf{85.4}    & 52.1      & \textbf{53.9}   & \textbf{93.8}      & 67.3 & \textbf{76.1}       & \textbf{61.8} \\ \Xhline{1px}
\end{tabular}
}
\end{table*}


\subsection{Implementation Details}
To address the absence of 3D human body part segmentation datasets based on LiDAR point clouds, we create a synthetic dataset of 1 million LiDAR human point cloud instances, named LiDARPart-Human, which uses the AMASS dataset for 3D human meshes and simulates LiDAR scans from various perspectives and distances (Figure.~\ref{fig:HBSeg}). These scans incorporate random occlusions and noise to reduce the gap between synthetic and real data. The SMPL mesh vertices, known for their ordered and regular structure, provide 24 human body part labels, but due to the sparsity of LiDAR point clouds, we simplify these to 9 main categories: head, left-arm, right-arm, up-body, low-body, upleft-leg, upright-leg, lowleft-leg, and lowright-leg. Each LiDAR point is automatically labeled with the nearest vertex's body part label. 
\subsection{Dataset and Evaluation Metrics}
Similar to LiDARFlow-Human (Section.~\ref{sec:hmflow}), we also establish a benchmark on LiDARPart-Human and will make it public.
As shown in Table.~\ref{tab:lidarpart_human}, the evaluation metric for LiDARPart-Human is the mean Intersection over Union (mIoU), which is the average of the IoUs calculated for each of the 9 human body parts.

\section{The Network Design Details for the Tokenizer}
\begin{figure}[t]
    \centering
    \includegraphics[width=1\columnwidth]{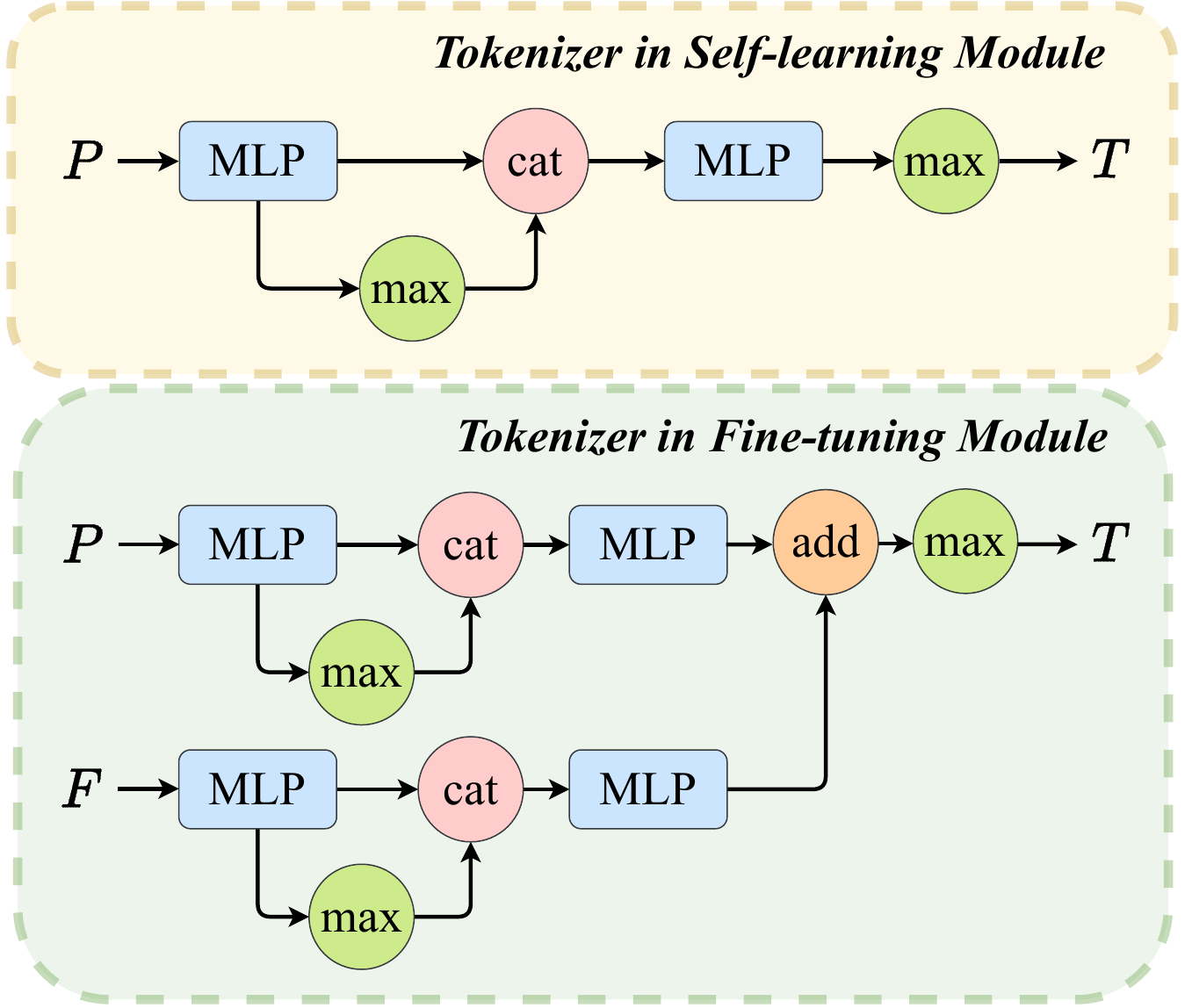}
    \caption{The network design details for the tokenizer. The primary distinction between the tokenizer in the fine-tuning module and that in the self-learning module lies in the integration of the motion flow features, denoted as $F$. } 
    \label{fig:tokenizer}
    \vspace{-1ex}
\end{figure}
As previously mentioned, in the self-learning module, the motion flow features $F$ are not fused with the part patches features $P$. This design prevents premature leakage of location information of masked tokens to the STEncoder.  
The network design details for the Tokenizer are illustrated in Figure.~\ref{fig:tokenizer}. 
During self-learning, each point of $P$ is mapped to a feature vector using several shared MLPs. Subsequently, max-pooled features are concatenated to each feature vector. These are then processed through several MLPs to expand their dimension to $C=384$. 
During fine-tuning, the same operation is applied to the motion flow $F$. The features of $P$ and $F$ are then fused through element-wise addition. Finally, a max-pooling layer is applied to derive the part token $T$.

\begin{table}[ht]
\centering
\caption{Comparative Analysis of Model Parameter Numbers in Transformer-Based Dynamic Point Cloud Methods.}
\begin{tabular}{l|cc}
\Xhline{1px}
                                                          & \multicolumn{2}{c}{Num of Params(M)$\downarrow$}            \\ \hline
                                                          & \multicolumn{1}{c|}{Self-learning} & Fine-tuning \\ \hline
P4Transformer~\cite{fan2021point}   & \multicolumn{1}{c|}{/}             & 40.37       \\ \hline
PST-Transformer~\cite{fan2022point} & \multicolumn{1}{c|}{/}             & 60.36       \\ \hline
PPTr~\cite{wen2022point}            & \multicolumn{1}{c|}{/}             & 120.7       \\ \hline
MaST-Pre~\cite{shen2023masked}      & \multicolumn{1}{c|}{140.76}        & 120.66      \\ \hline
UniPVU-Human                                              & \multicolumn{1}{c|}{\textbf{34.92}}         & \textbf{22.48}       \\ \Xhline{1px}
\end{tabular}
\label{tab:model_size}
\end{table}

\section{Supplementary Comparison Experiments on HuCenLife~\cite{xu2023human}}
For more comprehensive and extensive comparisons, we supplement our comparison experiments on HuCenLife~\cite{xu2023human} with methods specifically designed for modeling dynamic point cloud videos~\cite{fan2022pstnet,fan2021point,fan2022point,wen2022point,fan2021deep}. As can be seen from Table.~\ref{tab:sup_comp_exp}, although these methods perform better in categories that require modeling motion features for accurate recognition (Fitness, Sports, Bend-Over, Walk-Stand) than static point cloud methods, there is still a significant performance gap compared to our UniPVU-Human. This confirms the superiority of our method in capturing human motion representations. We also compared our method with self-learning approaches based on contrastive learning~\cite{shen2023pointcmp}. The experimental results demonstrate the superiority of our self-learning mechanism.

\section{Comparative Analysis of Model Parameter Numbers}
Transformer\cite{vaswani2017attention}-based methods~\cite{chen2023pointgpt,liu2023point,zhao2021point,guo2021pct} have achieved considerable performance in point cloud feature extraction. However, their large model size typically results in significant computational demands. As we can see from Table.~\ref{tab:model_size}, the parameter number of other transformer-based dynamic point cloud methods~\cite{fan2021point,fan2022point,wen2022point, shen2023masked}, are several times greater than that of our UniPVU-Human. Our model maintains a parameter number of twenty to thirty million in both the self-learning and fine-tuning stages, which is comparable to ResNet-50~\cite{he2016deep}. Therefore, our UniPVU-Human achieves better performance with fewer parameters, making it a lightweight and effective model well-suited for real-world applications.

\end{appendix}


\end{document}